\documentclass[10pt,twocolumn,letterpaper]{article}

\usepackage{iccv}
\usepackage{times}
\usepackage{epsfig}
\usepackage{graphicx}
\usepackage{amsmath}
\usepackage{amssymb}
\usepackage{subfigure}
\usepackage{tabularx}
\usepackage{color}
\usepackage{multirow}
\usepackage{colortbl}

\newcommand{\figref}[1]{Fig. \ref{#1}}
\newcommand{\tabref}[1]{Table \ref{#1}}
\newcommand{\equref}[1]{(\ref{#1})}

\definecolor{srcolor}{rgb}{1,0,0}

\makeatletter
\def\hlinewd#1{%
	\noalign{\ifnum0=`}\fi\hrule \@height #1 \futurelet
	\reserved@a\@xhline}

\usepackage[breaklinks=true,bookmarks=false]{hyperref}

\iccvfinalcopy 

\def\iccvPaperID{2581} 

\ificcvfinal\fi
\begin{document}

\title{DCTM: Discrete-Continuous Transformation Matching for Semantic Flow}

\author{Seungryong Kim\textsuperscript{1}, Dongbo Min\textsuperscript{2}, Stephen Lin\textsuperscript{3}, and Kwanghoon Sohn\textsuperscript{1}\\
	{\textsuperscript{1}Yonsei University \quad \textsuperscript{2}Chungnam National University \quad \textsuperscript{3}Microsoft Research}\\
	{\tt\small \{srkim89,khsohn\}@yonsei.ac.kr \quad dbmin@cnu.ac.kr \quad stevelin@microsoft.com}}

\maketitle

\begin{abstract}
	Techniques for dense semantic correspondence have provided limited ability to deal with the geometric variations that commonly exist between semantically similar images. While variations due to scale and rotation have been examined, there lack practical solutions for more complex deformations such as affine transformations because of the tremendous size of the associated solution space. To address this problem, we present a discrete-continuous transformation matching (DCTM) framework where dense affine transformation fields are inferred through a discrete label optimization in which the labels are iteratively updated via continuous regularization. In this way, our approach draws solutions from the continuous space of affine transformations in a manner that can be computed efficiently through constant-time edge-aware filtering and a proposed affine-varying CNN-based descriptor. Experimental results show that this model outperforms the state-of-the-art methods for dense semantic correspondence on various benchmarks.
\end{abstract}

\section{Introduction}\label{sec:1}
Establishing dense correspondences across \emph{semantically} similar images is essential for numerous tasks such as nonparametric scene parsing, scene recognition, image registration, semantic segmentation, and image editing \cite{HaCohen2011,Liu11,Liu11b}.

Unlike traditional dense correspondence for estimating depth \cite{Scharstein02} or optical flow \cite{Butler12,Sun10}, semantic correspondence estimation poses additional challenges due to intra-class appearance and shape variations among object instances, which can degrade matching by conventional approaches \cite{Liu11,Yang14}.
Recently, several methods have attempted to deal with the appearance differences using convolutional neural network (CNN) based descriptors because of their high invariance to appearance variations \cite{Long14,Choy16,Zhou16,Kim17}. 
However, geometric variations are considered in just a limited manner through constraint settings such as those used for depth or optical flow. Some methods solve for geometric variations such as scale or rotation \cite{Hassner12,Qiu14,Hur15}, but they consider only a discrete set of scales or rotations as possible solutions, and do not capture the non-rigid geometric deformations that commonly exist between semantically similar images.
\begin{figure}
	\centering
	\renewcommand{\thesubfigure}{}
	\subfigure[(a)]
	{\includegraphics[width=0.245\linewidth]{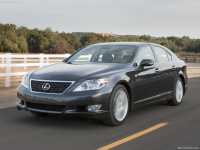}}\hfill
	\subfigure[(b)]
	{\includegraphics[width=0.245\linewidth]{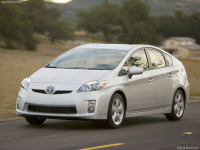}}\hfill
	\subfigure[(c)]
	{\includegraphics[width=0.245\linewidth]{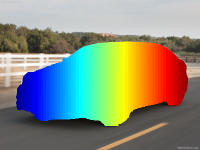}}\hfill
	\subfigure[(d)]
	{\includegraphics[width=0.245\linewidth]{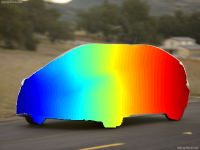}}\hfill
	\vspace{-6pt}
	\subfigure[(e)]
	{\includegraphics[width=0.245\linewidth]{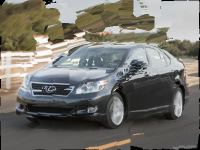}}\hfill
	\subfigure[(f)]
	{\includegraphics[width=0.245\linewidth]{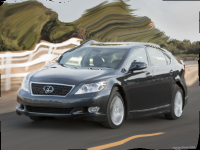}}\hfill
	\subfigure[(g)]
	{\includegraphics[width=0.245\linewidth]{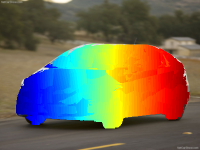}}\hfill
	\subfigure[(h)]
	{\includegraphics[width=0.245\linewidth]{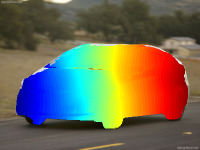}}\hfill
	\vspace{-1pt}
	\caption{Visualization of our DCTM results: (a) source image, (b) target image, (c), (d) ground truth correspondences, (e), (f), (g), (h) warped images and correspondences after discrete and continuous optimization, respectively. For images undergoing non-rigid deformations, our DCTM estimates reliable correspondences by iteratively optimizing the label space via continuous regularization.}\label{img:1}\vspace{-10pt}
\end{figure}

It has been shown that these non-rigid image deformations can be locally well approximated by affine transformations \cite{Schaefer06,Lin11,Lin12}. To estimate dense affine transformation fields, a possible approach is to discretize the space of affine transformations and find a labeling solution. However, the higher-dimensional search space for affine transformations makes discrete global optimization algorithms such as graph cut \cite{Boykov01} and belief propagation \cite{Shekhovtsov07,Szeliski08} computationally infeasible. For more efficient optimization over large label spaces, the PatchMatch Filter (PMF) \cite{Lu13} integrates constant-time edge-aware filtering (EAF) \cite{Rhemann11,Lu12} with PatchMatch-based randomized search \cite{Barnes10}. PMF is leveraged for dense semantic correspondence in DAISY Filter Flow (DFF) \cite{Yang14}, which finds labels for displacement fields as well as for scale and rotation. Extending DFF to affine transformations would be challenging though. One reason is that its efficient technique for computing DAISY features \cite{Tola10} at pre-determined scales and rotations cannot be applied for affine transformations. Another reason is that, as shown in \cite{Li15,Hur15}, the weak implicit smoothing embedded in PMF makes it more susceptible to erroneous local minima, and this problem may be magnified in the higher-dimensional affine transformation space. Explicit smoothing models have been adopted to alleviate this problem in the context of stereo matching \cite{Lin13,Besse14}, but were designed specifically for depth regularization.

In this paper, we introduce an effective method for estimating dense affine transformation fields between semantically similar images, as shown in \figref{img:1}. The key idea is to couple a discrete local labeling optimization with a continuous global regularization that updates the discrete candidate labels. An affine transformation field is efficiently inferred in a filter-based discrete labeling scheme inspired by PMF, and then
the discrete affine transformation field is
globally regularized in a moving least squares (MLS) manner
\cite{Schaefer06}.
These two steps are iterated in alternation until convergence. Through the synergy of the discrete local labeling and continuous global regularization, our method yields {\em continuous} solutions from the space of affine transformations, rather than selecting from a pre-defined, finite set of discrete samples. We show that this continuous regularization additionally overcomes the aforementioned implicit smoothness problem in PMF.

Moreover, we model the effects of affine transformations directly within the state-of-the-art fully convolutional self-similarity (FCSS) descriptor \cite{Kim17}, which leads to significant improvements in processing speed over computing descriptors on various affine transformations of the image. Experimental results show that the presented model outperforms the latest methods for dense semantic correspondence on several benchmarks, including that of Taniai \emph{et al.}~\cite{Taniai16}, Proposal Flow~\cite{Ham16}, and PASCAL~\cite{Chen14}.


\section{Related Work}\label{sec:2}
\paragraph{Dense Semantic Flow}\label{sec:21}
Most conventional techniques for dense semantic correspondence have employed handcrafted features such as SIFT \cite{Lowe04} or DAISY \cite{Tola10}. To improve matching quality, they have focused on optimization.
Liu \emph{et al.} \cite{Liu11} pioneered the idea of dense correspondence across different scenes, and proposed SIFT Flow which is based on hierarchical dual-layer belief propagation.
Inspired by this, Kim \emph{et al.} \cite{Kim13} proposed the deformable spatial pyramid (DSP) which performs multi-scale regularization with a hierarchical graph. Among other methods are those that take an exemplar-LDA approach \cite{Bristow15}, employ joint image set alignment \cite{Zhou15}, or jointly solve for cosegmentation \cite{Taniai16}.

Recently, CNN-based descriptors have been used to establish dense semantic correspondences. 
Zhou \emph{et al.} \cite{Zhou16} proposed a deep network that exploits cycle-consistency with a 3D CAD model \cite{ShapeNet} as a supervisory signal. Choy \emph{et al.} \cite{Choy16} proposed the universal correspondence network (UCN) based on fully convolutional feature learning. Most recently, Kim \emph{et al.} \cite{Kim17} proposed the FCSS descriptor that formulates local self-similarity (LSS) \cite{Schechtman07} within a fully convolutional network. Because of its LSS-based structure, FCSS is inherently insensitive to intra-class appearance variations while maintaining precise localization ability. However, none of these methods is able to handle non-rigid geometric variations.

Several methods aim to alleviate geometric variations through extensions of SIFT Flow, including scale-less SIFT Flow (SLS) \cite{Hassner12}, scale-space SIFT Flow (SSF) \cite{Qiu14}, and generalized DSP (GDSP) \cite{Hur15}.
However, these techniques have a critical practical limitation that their computation increases linearly with the search space size.
A generalized PatchMatch algorithm \cite{Barnes10} was proposed for efficient matching that leverages a randomized search scheme. This was utilized by HaCohen \emph{et al.} \cite{HaCohen2011} in a non-rigid dense correspondence (NRDC) algorithm, but employs weak matching evidence that cannot guarantee reliable performance. Geometric invariance to scale and rotation is provided by DFF \cite{Yang14}, but its implicit smoothing model which relies on randomized sampling and propagation of good estimates in the direct neighborhood often induces mismatches.
A segmentation-aware approach \cite{Trulls13} was proposed to provide geometric robustness for descriptors, \emph{e.g.}, SIFT \cite{Lowe04}, but can have a negative effect on the discriminative power of the descriptor. Recently, Ham \emph{et al.} \cite{Ham16} presented the Proposal Flow (PF) algorithm to estimate correspondences using object proposals. While these aforementioned techniques provide some amount of geometric invariance, none of them can deal with
affine transformations across images, which are a frequent occurrence in dense semantic correspondence.\vspace{-10pt}

\paragraph{Image Manipulation}\label{sec:23}
A possible approach for estimating dense affine transformation fields is to interpolate sparsely matched points using a method, including thin plate splines (TPS) \cite{Bookstein89}, motion coherence \cite{Yuille88}, coherence point drift \cite{Myronenko07}, or smoothly varying affine stitching \cite{Lin11}. MLS is also a scattered point interpolation technique first introduced in \cite{Lancaster81} to reconstruct a continuous function from a set of point samples by incorporating spatially-weighted least squares. MLS has been successfully used in applications such as image deformation \cite{Schaefer06}, surface reconstruction \cite{Fleishman05}, image super-resolution and denoising \cite{Bose06}, and color transfer \cite{Hwang14}. Inspired by the MLS concept, our method utilizes it to regularize estimated affine transformation fields, but with a different weight function and an efficient computational scheme.

More related to our work is the method of Lin \emph{et al.} \cite{Lin12}, which jointly estimates correspondence and relative patch orientation for descriptors.
However, it is formulated with pre-computed sparse correspondences and also requires considerable computation to solve a complex non-linear optimization. By contrast, our method adopts dense descriptors that can be evaluated efficiently for any affine transformation, and employs quadratic continuous optimization to rapidly infer dense affine transformation fields.

\section{Method}\label{sec:3}
\subsection{Problem Formulation and Model}\label{sec:31}
Given a pair of images $I$ and $I'$,
the objective of dense correspondence estimation is to establish a correspondence $i'$ for each pixel $i=[i_\mathbf{x},i_\mathbf{y}]$.
Unlike conventional dense correspondence settings
for estimating depth \cite{Scharstein02}, optical flow \cite{Butler12,Sun10}, or similarity transformations \cite{Yang14,Hur15},
our objective is to infer a field of affine transformations, each represented by a $2\times3$ matrix
\begin{equation}
	\mathbf{T}_i = \left[ {\begin{array}{*{20}{c}}
			{\mathbf{T}_{i,\mathbf{x}}}\\
			{\mathbf{T}_{i,\mathbf{y}}}
		\end{array}} \right]
	\end{equation}
	that maps pixel $i$ to ${i'} = \mathbf{T}_{i}\mathbf{i}$, where $\mathbf{i}$ is pixel $i$ represented in homogeneous coordinates such that $\mathbf{i}=[i,1]^T$.
	
	In this work, we solve for affine transformations that may lie anywhere in the continuous solution space. This is made possible by formulating the inference of dense affine transformation fields as a discrete optimization problem with continuous regularization. This optimization seeks to minimize an energy of the form
	\begin{equation}\label{equ:energy-func}
		\begin{split}
			E(\mathbf{T}) = E_{data}(\mathbf{T}) + \lambda E_{smooth}(\mathbf{T}),
		\end{split}
	\end{equation}
	consisting of a data term that accounts for matching evidence between descriptors
	and a smoothness term that favors similar affine transformations among adjacent pixels with a balancing parameter $\lambda$.
	
	Our data term is defined as follows:
	\begin{equation}\label{equ:energy-func-data}
		E_{data}(\mathbf{T}) = \sum\limits_i {\sum\limits_{j \in {\mathcal{N}_i}} {{\omega^I_{ij}}\min (\|\mathcal{D}_j - \mathcal{D}'_{j'}(\mathbf{T}_i)\|_1,\tau )}}.
	\end{equation}
	It is designed to estimate the affine transformation $\mathbf{T}_i$ by aggregating the matching costs of descriptors between neighboring pixels $j$ and transformed pixels $j' = {\mathbf{T}_i}\mathbf{j}$ within a local aggregation window ${\mathcal{N}_i}$. A truncation threshold $\tau$ is used to deal with outliers and occlusions. It should be noted that aggregated data terms have been popularly used in stereo \cite{Scharstein02} and optical flow \cite{Li15}.
	For dense semantic correspondence,
	several methods have employed aggregated data terms;
	however, they often produce undesirable results across object boundaries due to uniform weights that ignore image structure \cite{Kim13,Hur15}, or fail to deal with geometric distortions like affine transformations as they rely on a regular grid structure for local aggregation windows \cite{Yang14}. By contrast, the proposed method adaptively aggregates matching costs using edge-preserving bilateral weights ${\omega^I_{ij}}$ as in \cite{Tomasi98,He10} on a geometrically-variant grid structure in order to produce spatially smooth yet discontinuity-preserving labeling results even under affine transformations.
	
	Our smoothness term is defined as follows to regularize affine transformation fields $\mathbf{T}_i$ within a local neighborhood:
	\begin{equation}\label{equ:energy-func-smooth}
		E_{smooth}(\mathbf{T}) = \sum\limits_i {\sum\limits_{j \in {\mathcal{M}_i}} {{\upsilon^I_{ij}}\|{\mathbf{T}_i}\mathbf{j} - {\mathbf{T}_j}\mathbf{j}\|^2}}.
	\end{equation}
	When the affine transformation $\mathbf{T}$ is constrained to $[\mathbf{I}_{2\times 2},\mathbf{u}]$ with $\mathbf{u} = [u_{\mathbf{x}},u_{\mathbf{y}}]^T$ and ${\mathcal{M}_i}$ is the 4-neighborhood, this smoothness term becomes the first order derivative of the optical flow vector as in many conventional methods \cite{Liu11,Min14}. However, non-rigid deformations occur with high frequency in semantic correspondence, and such a basic constraint is inadequate for modeling the smoothness of affine transformation fields. Our smoothness term is formulated to address this by regularizing estimated affine transformations $\mathbf{T}_i$ in a moving least squares manner \cite{Schaefer06} within local neighborhood ${\mathcal{M}_i}$.
	We define the smoothness constraint of affine transformation fields by fitting $\mathbf{T}_i$ based on the affine flow fields of neighboring pixels ${\mathbf{T}_j}\mathbf{j}$. Unlike conventional moving least square solvers \cite{Schaefer06},
	our smoothness term incorporates edge-preserving bilateral weights ${\upsilon^I_{ij}}$ as in \cite{Tomasi98,He10} for image structure-aware regularization.
	
	Minimizing the energy in \equref{equ:energy-func} is a non-convex optimization problem defined over an infinite continuous solution space. A similar issue exists for optical flow estimation \cite{Brox09,Xu10,Revaud15}. To minimize the non-convex energy function, several techniques such as a hybrid method with descriptor matching \cite{Brox09,Revaud15} and a coarse-to-fine scheme \cite{Xu10} have been used, but they are tailored to optical flow estimation and have exhibited limited performance. We instead use a penalty decomposition scheme to alternately solve for the discrete and continuous affine transformation fields. An efficient filter-based discrete optimization technique is used to locally estimate discrete
	affine transformations in a manner similar to PMF \cite{Lu13}. The weakness of the implicit
	smoothing in the discrete local optimization is overcome by regularizing the affine transformation fields through global optimization in the continuous space. 
	This alternating optimization is repeated until convergence.
	Furthermore, to acquire matching evidence for semantic
	correspondence under spatially-varying affine fields, we extend the FCSS descriptor \cite{Kim17} to model affine variations.
	\begin{figure}
		\centering
		\renewcommand{\thesubfigure}{}
		\subfigure[(a) FCSS \cite{Kim17}]
		{\includegraphics[width=0.37\linewidth]{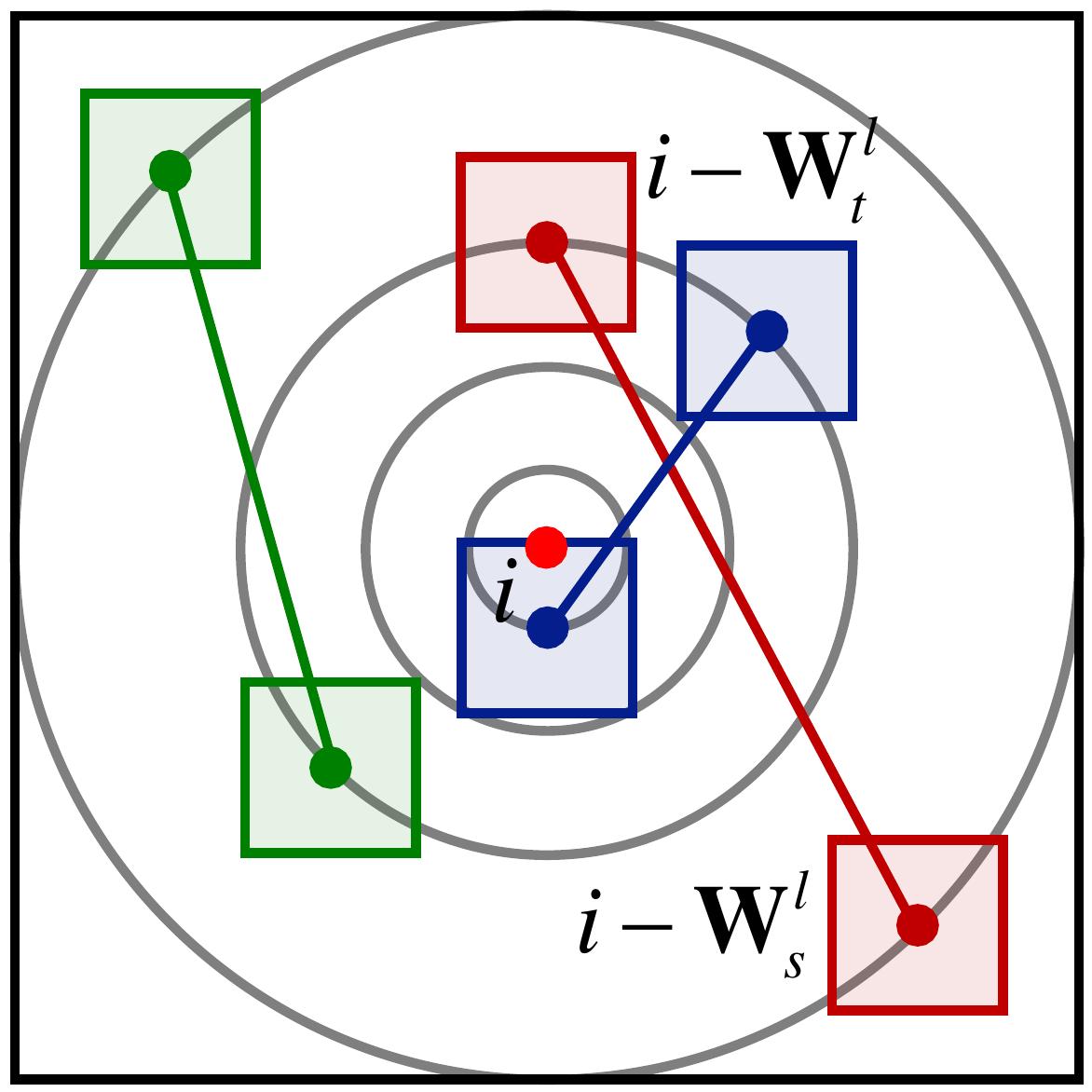}}
		\subfigure[(b) Affine-FCSS]
		{\includegraphics[width=0.37\linewidth]{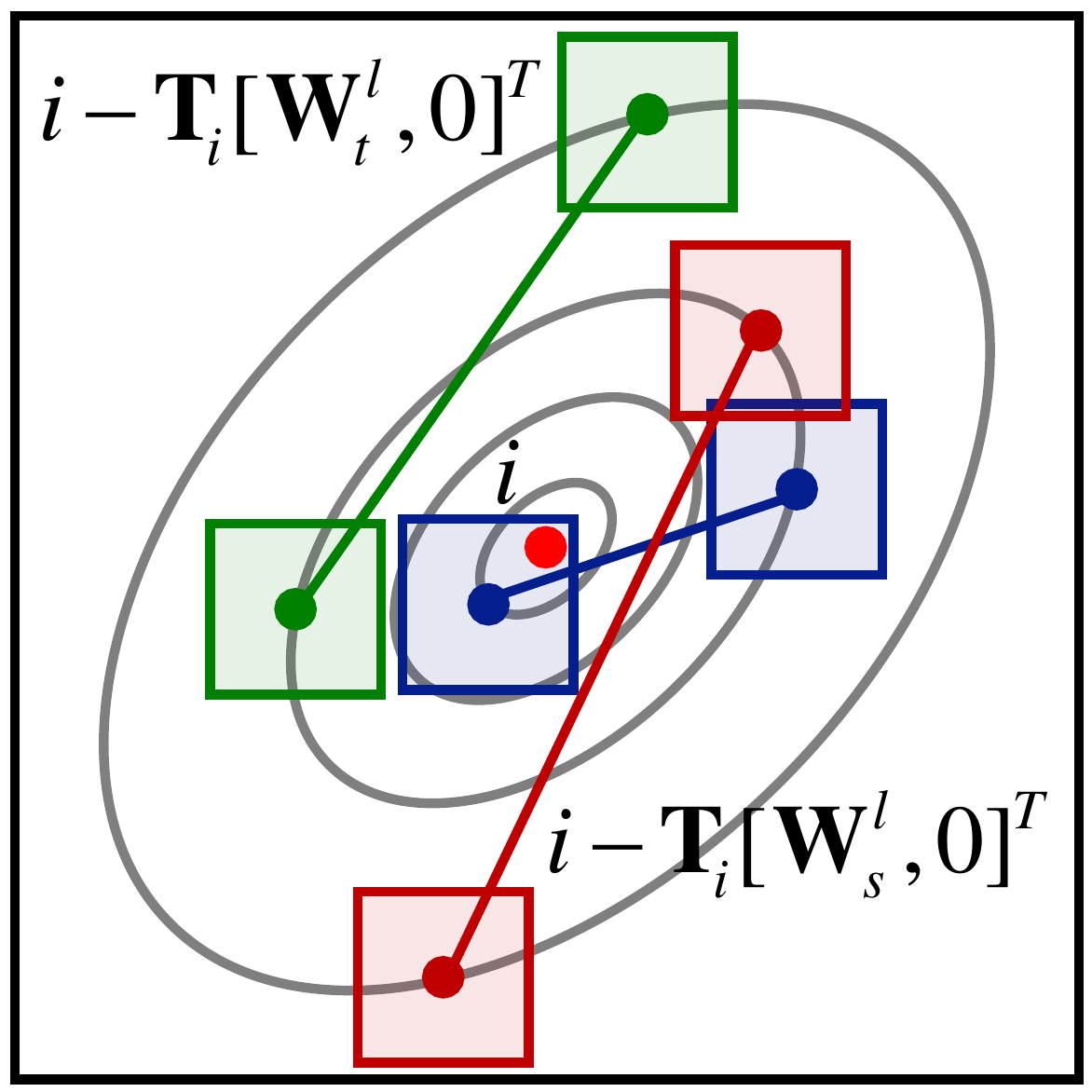}}\\
		\vspace{-1pt}
		\caption{Illustration of (a) FCSS descriptor \cite{Kim17} and (b) affine-FCSS descriptor. Within a support window, sampling patterns $\mathbf{W}^l_s$ and $\mathbf{W}^l_t$ are transformed according to affine fields $\mathbf{T}_i$.}\label{img:2}\vspace{-10pt}
	\end{figure}
	\subsection{Affine-FCSS Descriptor}\label{sec:32}
	To estimate a matching cost, a dense descriptor ${\mathcal{D}_i}$ is extracted over the local support window of each image point $I_{i}$. For this we employ the state-of-the-art FCSS descriptor \cite{Kim17} for dense semantic correspondence, which formulates LSS \cite{Schechtman07} within a fully convolutional network in a manner
	where the patch sampling patterns and self-similarity measure are both learned.
	Formally, FCSS can be described as a vector of feature values $\mathcal{D}_{i} = { \bigcup_{l}} \mathcal{D}^l_{i}$ for $l \in \{1,...,L\}$ with the maximum number of sampling patterns $L$, where the feature values are computed as
	\begin{equation}\label{equ:fcss-descriptor}
		\mathcal{D}^l_{i} = \mathrm{exp} ( - \mathcal{S} (i-\mathbf{W}^l_s,i-\mathbf{W}^l_t)/\mathbf{W}_\sigma ).
	\end{equation}
	$\mathcal{S}(\cdot,\cdot)$ represents the self-similarity between two convolutional activations taken from
	a sampling pattern around center pixel $i$, and can be expressed as
	\begin{equation}\label{equ:css-efficient}
		\mathcal{S}(i-\mathbf{W}^l_s,i-\mathbf{W}^l_t) = \|\mathcal{F}
		(\mathbf{A}_i;\mathbf{W}^l_s) - \mathcal{F}
		(\mathbf{A}_i;\mathbf{W}^l_t)\|^2,
	\end{equation}
	where $\mathcal{F}(\mathbf{A}_i;\mathbf{W}^l_s) = \mathbf{A}_{i-\mathbf{W}^l_s}$ and $\mathcal{F}(\mathbf{A}_i;\mathbf{W}^l_t) = \mathbf{A}_{i-\mathbf{W}^l_t}$, $\mathbf{W}^l_{s} = [W^{l}_{s,\mathbf{x}},W^{l}_{s,\mathbf{y}}]$ and $\mathbf{W}^l_{t} = [W^{l}_{t,\mathbf{x}},W^{l}_{t,\mathbf{y}}]$ compose the $l$-th learned sampling pattern, and $\mathbf{A}_i$
	is the convolutional activation through feed-forward process $\mathcal{F} (I_i;\mathbf{W}_c)$ for $I_i$ with network weights $\mathbf{W}_c$. The network parameters $\mathbf{W}_c$, $\mathbf{W}_s$, $\mathbf{W}_t$, and $\mathbf{W}_\sigma$
	are learned in an end-to-end manner to provide optimal correspondence performance.
	
	The FCSS descriptor provides high invariance to appearance variations, but it inherently cannot deal with geometric variations due to its pre-defined sampling patterns for all pixels in an image. Furthermore, although its computation is efficient, FCSS cannot in practice be evaluated exhaustively over all the affine candidates during optimization. To alleviate these limitations, we extend the FCSS descriptor to adapt to affine transformation fields.
	This is accomplished by reformulating the sampling patterns so that they account for the affine transformations. To expedite this computation, we first compute $\mathbf{A}_i$ over the entire image domain by passing it through the network. An FCSS descriptor $\mathcal{D}_i(\mathbf{T}_i)$ transformed under
	an affine field $\mathbf{T}_i$ can then be built by computing self-similarity on transformed sampling patterns
	\begin{equation}\label{equ:css-efficient}
		\|\mathcal{F}(\mathbf{A}_i;\mathbf{T}_i[\mathbf{W}^{l}_s,0]^T) - \mathcal{F}(\mathbf{A}_i;\mathbf{T}_i[\mathbf{W}^{l}_t,0]^T)\|^2.
	\end{equation}
	With this approach, repeated computation of convolutional activations over different affine transformations of the image is avoided.
	The affine transformation is efficiently inferred in a discrete optimization described in the following section. Differences between the FCSS descriptor and the affine-FCSS descriptor are illustrated in \figref{img:2}.
	\begin{figure}
		\centering
		\renewcommand{\thesubfigure}{}
		\subfigure[]
		{\includegraphics[width=1\linewidth]{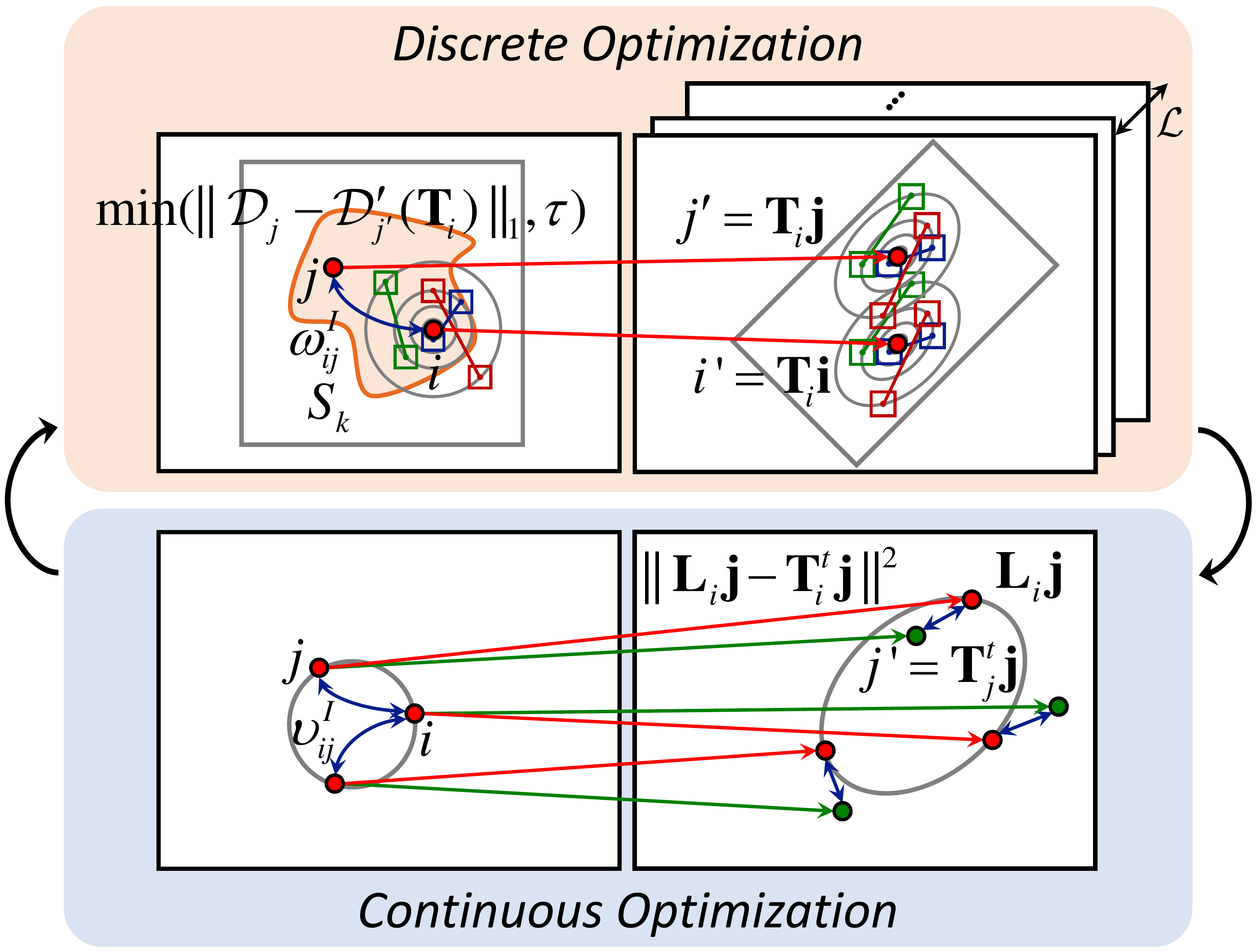}}\\
		\vspace{-10pt}
		\caption{Our DCTM method consists of discrete optimization and continuous optimization. Our DCTM method differs from the conventional PMF \cite{Lu13} by alternately optimizing the discrete label space and performing the continuous regularization.
		}\label{img:3}\vspace{-10pt}
	\end{figure}
	\begin{figure*}
		\centering
		\renewcommand{\thesubfigure}{}
		\subfigure[]
		{\includegraphics[width=0.122\linewidth]{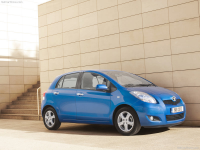}}\hfill
		\subfigure[]
		{\includegraphics[width=0.122\linewidth]{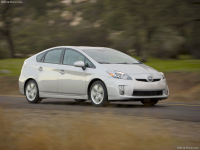}}\hfill
		\subfigure[]
		{\includegraphics[width=0.122\linewidth]{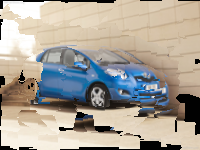}}\hfill
		\subfigure[]
		{\includegraphics[width=0.122\linewidth]{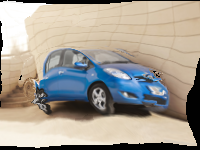}}\hfill
		\subfigure[]
		{\includegraphics[width=0.122\linewidth]{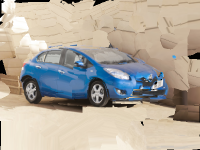}}\hfill
		\subfigure[]
		{\includegraphics[width=0.122\linewidth]{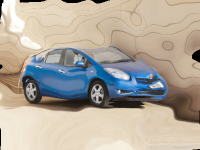}}\hfill
		\subfigure[]
		{\includegraphics[width=0.122\linewidth]{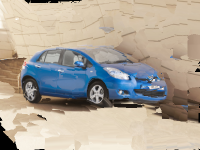}}\hfill
		\subfigure[]
		{\includegraphics[width=0.122\linewidth]{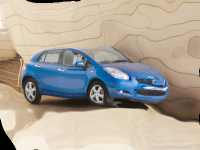}}\hfill
		\vspace{-21.5pt}
		\subfigure[(a)]
		{\includegraphics[width=0.122\linewidth]{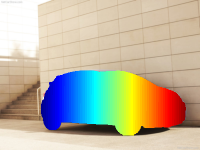}}\hfill
		\subfigure[(b)]
		{\includegraphics[width=0.122\linewidth]{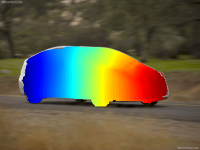}}\hfill
		\subfigure[(c)]
		{\includegraphics[width=0.122\linewidth]{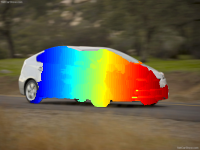}}\hfill
		\subfigure[(d)]
		{\includegraphics[width=0.122\linewidth]{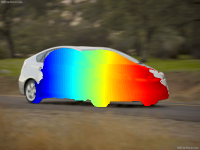}}\hfill
		\subfigure[(e)]
		{\includegraphics[width=0.122\linewidth]{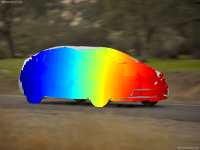}}\hfill
		\subfigure[(f)]
		{\includegraphics[width=0.122\linewidth]{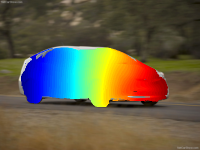}}\hfill
		\subfigure[(g)]
		{\includegraphics[width=0.122\linewidth]{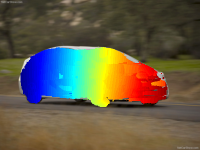}}\hfill
		\subfigure[(h)]
		{\includegraphics[width=0.122\linewidth]{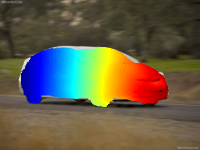}}\hfill
		\vspace{-1pt}
		\caption{DCTM convergence: (a) Source image; (b) Target image; Iterative evolution of warped images (c), (e), (g) after discrete optimization and (d), (f), (h) after continuous optimization. Our DCTM optimizes the label space with continuous regularization during the iterations, which facilitates convergence and boosts matching performance.}\label{img:4}\vspace{-10pt}
	\end{figure*}
	\subsection{Solution}\label{sec:34}
	Since affine transformation fields are defined in an infinite label space, minimizing our energy function $E(\mathbf{T})$ directly is infeasible. Through fine-scale discretization of this space, affine transformation fields could be estimated through discrete global optimization, but at a tremendous computational cost.
	To address this issue,
	we introduce an auxiliary affine field $\mathbf{L}$ to decouple our data and regularization terms, and approximate the original minimization
	problem as the following auxiliary energy formulation:
	\begin{equation}\label{equ:energy-func-aux}
		\begin{split}
			E_{\mathrm{aux}}&(\mathbf{T},\mathbf{L}) = \sum\limits_i {\sum\limits_{j \in {\mathcal{N}_i}} {{\omega^I_{ij}}\min (\|\mathcal{D}_j - \mathcal{D}'_{j'}(\mathbf{T}_i)\|_1,\tau )}} \\
			&+ \mu \sum\limits_i {\|\mathbf{L}_i-\mathbf{T}_i\|^2}
			+ \lambda \sum\limits_i {\sum\limits_{j \in {\mathcal{M}_i}} {{\upsilon^I_{ij}}\|{\mathbf{L}_i}\mathbf{j} - {\mathbf{T}_j}\mathbf{j}\|^2}}.
		\end{split}
	\end{equation}
	
	Since this energy function is based on two affine transformations, $\mathbf{L}$ and $\mathbf{T}$, we employ alternating minimization to solve for them and boost matching performance in a synergistic manner.
	We split the optimization of $E_{\mathrm{aux}}(\mathbf{L},\mathbf{T})$ into two sub-problems,
	namely a discrete local optimization problem with respect to
	$\mathbf{T}$ and a continuous global optimization problem with respect to
	$\mathbf{L}$. Increasing $\mu$ through the iterations drives the affine fields $\mathbf{T}$ and $\mathbf{L}$ together and eventually results in $\mathop {\lim }\nolimits_{\mu  \to \infty }E_{\mathrm{aux}} \approx E$. \vspace{-10pt}
	
	\paragraph{Discrete Optimization}
	To infer the discrete affine transformation field $\mathbf{T}^{t}$ with $\mathbf{L}^{t-1}$ being fixed at the
	$t$-th iteration, we first discretize the continuous parameter space and then solve the problem through filter-based label inference. For discrete affine transformation candidates $\mathbf{T} \in \mathcal{L}$, the matching cost between FCSS
	descriptors $\mathcal{D}_j$ and $\mathcal{D}'_{j'}(\mathbf{T})$ is
	first measured as
	\begin{equation}
		C_j(\mathbf{T}) = \min (\|\mathcal{D}_j - \mathcal{D}'_{j'}(\mathbf{T})\|_1,\tau ),
	\end{equation}
	where $\mathcal{D}'_{j'}(\mathbf{T})$ is the affine-FCSS descriptor with respect to $\mathbf{T}$. This yields an affine-invariant matching cost. Furthermore, since $j'$ varies according to affine fields
	such that $j' = {\mathbf{T}}\mathbf{j}$,
	affine-varying regular grids can be used when aggregating matching costs, thus enabling affine-invariant cost aggregation.
	To aggregate the raw matching costs,
	we apply EAF on $C_i(\mathbf{T})$ such that
	\begin{equation}
		\bar{C}_i(\mathbf{T}) = \sum\nolimits_{j \in {\mathcal{N}_i}}{\omega ^I_{ij} C_j(\mathbf{T})},
	\end{equation}
	where $\omega^I_{ij}$ is the normalized adaptive weight of a support pixel $j$, which can be defined in various ways with respect to the structures of the image $I$ \cite{Tomasi98,Gastal11,He10}.
	
	In determining the affine field $\mathbf{T}$, the matching costs are also augmented by the previously estimated affine transformation field $\mathbf{L}^{t-1}_i$ such that
	\begin{equation}
		G_i(\mathbf{T}) = \mu{\|\mathbf{T}-\mathbf{L}^{t-1}_i\|^2} + \lambda \sum\limits_{j \in {\mathcal{M}_i}} {{\upsilon^I_{ij}}\|{\mathbf{T}}\mathbf{j} - {\mathbf{L}^{t-1}_i}\mathbf{j}\|^2}.
	\end{equation}
	Since $\|{\mathbf{T}}\mathbf{j} - {\mathbf{L}^{t-1}_i}\mathbf{j}\|^2 = \|({\mathbf{T}} - {\mathbf{L}^{t-1}_i})\mathbf{j}\|^2$ and ${\mathbf{T}} - {\mathbf{L}^{t-1}_i}$ is independent to pixel $j$ within the support window,
	$G_i(\mathbf{T})$ can be efficiently computed by using constant-time EAF, as described in detail in the supplementary material.
	
	The resultant label at the $t$-th iteration is
	determined with a winner-takes-all (WTA) scheme: 
	\begin{equation}\label{equ:sol-T}
		\mathbf{T}^{t}_i = \mathop{\mathrm{argmin}}\nolimits_{\mathbf{T}
			\in \mathcal{L}} \{\bar{C}_i(\mathbf{T}) + G_i(\mathbf{T})\}.
	\end{equation}
	\vspace{-25pt}
	\begin{table}
		\begin{center}
			\begin{tabularx}{\linewidth}{p{1.5mm} p{0.5mm}| p{0.5mm}| p{0.5mm}| p{160mm}}
				\hlinewd{1.0pt}
				\multicolumn{5}{ p{164mm} }{{\bf Algorithm 1}: DCTM Framework}\\
				\hlinewd{0.8pt}
				\multicolumn{5}{ p{164mm} }{{\bf Input}: images $I$, $I'$, FCSS network parameter $\mathbf{W}$}\\
				\multicolumn{5}{ p{164mm} }{{\bf Output}: dense affine transformation fields $\mathbf{T}$}\\
				\multicolumn{5}{ p{164mm} }{{\bf Parameters}: number of segments $K$, pyramid levels $F$}\\
				~&\multicolumn{4}{ p{163mm} }{{\bf $/*$ \emph{Initialization} $*/$}}\\
				$\mathbf{1:}$&\multicolumn{4}{ p{163mm} }{Partition $I$ into a set of disjoint $K$ segments $\{S_k\}$}\\
				$\mathbf{2:}$&\multicolumn{4}{ p{163mm} }{Initialize affine fields as $\mathbf{T}_i=[\mathbf{I}_{2\times2},\mathbf{0}_{2\times1}]$}\\
				~&\multicolumn{4}{ p{163mm} }{{\bf for} $f = 1:F$ {\bf do }}\\
				$\mathbf{3:}$&~&\multicolumn{3}{ p{162mm} }{Build convolution activations $\mathbf{A}^f$, $\mathbf{A}'^f$ for $I^f$, $I'^f$}\\
				$\mathbf{4:}$&~&\multicolumn{3}{ p{162mm} }{Initialize affine fields $\mathbf{T}^f_i = \mathbf{L}^{f-1}_i$ when $f > 2$}\\
				~&~&\multicolumn{3}{ p{162mm} }{{\bf while} not converged {\bf do }}\\
				~&~&~&\multicolumn{2}{ p{161mm} }{{\bf $/*$ \emph{Discrete Optimization} $*/$}}\\
				$\mathbf{5:}$&~&~&\multicolumn{2}{ p{161mm} }{Initialize affine fields $\mathbf{T}^{t}_i = \mathbf{L}^{t-1}_i$}\\
				~&~&~&\multicolumn{2}{ p{161mm} }{{\bf for} $k = 1:K$ {\bf do }}\\
				~&~&~&~&$\;${{\bf $/*$ \emph{Propagation} $*/$}}\\
				$\mathbf{6:}$&~&~&~&$\;$For $S_k$, construct affine candidates \par $\;\;\;$$\mathbf{T}\in \mathcal{L}_{p}$ from neighboring segments\\
				$\mathbf{7:}$&~&~&~&$\;$Build cost volumes $\bar{C}_i(\mathbf{T})$ and $G_i(\mathbf{T})$\\
				$\mathbf{8:}$&~&~&~&$\;$Determine $\mathbf{T}^{t}_i$ using \equref{equ:sol-T}\\
				~&~&~&~&$\;${{\bf $/*$ \emph{Random Search} $*/$}}\\
				$\mathbf{9:}$&~&~&~&$\;$Construct affine candidates $\mathbf{T} \in \mathcal{L}_{r}$ \par $\;\;\;$from randomly sampled affine fields\\
				$\mathbf{10:}$&~&~&~&$\;$Determine $\mathbf{T}^{t}_i$ by Step $\mathbf{7}$-$\mathbf{8}$\\
				~&~&~&\multicolumn{2}{ p{161mm} }{{\bf end for }}\\
				~&~&~&\multicolumn{2}{ p{161mm} }{{\bf $/*$ \emph{Continuous Optimization} $*/$}}\\
				$\mathbf{11:}$&~&~&\multicolumn{2}{ p{161mm} }{Estimate affine fields $\mathbf{L}^{t}_i$ from $\mathbf{T}^{t}_i$ using \equref{equ:sol-L}}\\
				~&~&\multicolumn{3}{ p{162mm} }{{\bf end while }}\\
				~&\multicolumn{4}{ p{163mm} }{{\bf end for }}\\
				\hlinewd{1.0pt}
			\end{tabularx}
		\end{center}\label{alg:1}\vspace{-20pt}
	\end{table}
	\paragraph{Continuous Optimization}
	To solve the continuous affine transformation field $\mathbf{L}^{t}$ with $\mathbf{T}^{t}$ being fixed,
	we formulate the problem as an image warping minimization:
	\begin{equation}
		\sum\limits_i \left(
		\mu {\|\mathbf{L}_i-\mathbf{T}^{t}_i\|^2} + \lambda{\sum\limits_{j \in {\mathcal{M}_i}} {{\upsilon^I_{ij}}\|{\mathbf{L}_i}\mathbf{j}-{\mathbf{T}^{t}_j}\mathbf{j}\|^2}}\right).
	\end{equation}
	
	Since this involves solving spatially-varying weighted least squares at each pixel $i$, the computational burden inevitably increases when considering non-local neighborhoods $\mathcal{M}_i$. To
	expedite this, existing MLS solvers adopted grid-based sampling
	\cite{Schaefer06} at the cost of quantization errors or parallel processing
	\cite{Hwang14} with additional hardware. 
	In contrast, our method optimizes the objective with a sparse matrix
	solver, yielding a substantial runtime gain. Since the $\mathbf{L}_i \mathbf{j}$ term can be
	formulated in the $\mathbf{x}$- and $\mathbf{y}$-directions
	separatively, $[\mathbf{L}_{i,\mathbf{x}}
	\mathbf{j},\mathbf{L}_{i,\mathbf{y}} \mathbf{j}]^T$, we decompose
	the objective into two separable energy functions. For the $\mathbf{x}$-direction, the energy function can be represented as
	\begin{equation}
		\begin{split}
			\sum\limits_i \left(
			\mu {\|\mathbf{L}_{i,\mathbf{x}}-\mathbf{T}^{t}_{i,\mathbf{x}}\|^2} + \lambda{\sum\limits_{j \in {\mathcal{M}_i}} \upsilon^I_{ij}\|\mathbf{L}_{i,\mathbf{x}} \mathbf{j}-{\mathbf{T}^{t}_{j,\mathbf{x}}}\mathbf{j}\|^2}\right).
		\end{split}
	\end{equation}
	By setting the gradient of this objective with respect to
	$\mathbf{L}_{\mathbf{x},i}$ to zero, the minimizer
	$\mathbf{L}^{t}_{i,\mathbf{x}}$ is obtained by solving a linear system based on a large sparse matrix:
	\begin{equation}\label{equ:sol-L}
		(\mu/\lambda\mathbf{I}+\mathbf{U})\mathbf{L}^{t}_{\mathbf{x}} = (\mu/\lambda\mathbf{I}+\mathbf{K}) \mathbf{T}^{t}_{\mathbf{x}},
	\end{equation}
	where $\mathbf{I}$ denotes a $3N\times3N$ identity matrix with $N$ denoting the number of pixels in image $I$.
	$\mathbf{L}^{t}_{\mathbf{x}}$ and $\mathbf{T}^{t}_{\mathbf{x}}$ denote $3N\times 1$ column vectors containing $\mathbf{L}^{t}_{i,\mathbf{x}}$
	and $\mathbf{T}^{t}_{i,\mathbf{x}}$, respectively. $\mathbf{U}$ and $\mathbf{K}$ denote matrices defined as
	\begin{equation}
		\mathbf{U} = \left[ {\begin{array}{*{20}{c}}
				\psi(\mathbf{V}X^2)&\psi(\mathbf{V}XY)&\psi(\mathbf{V}X)\\
				\psi(\mathbf{V}XY)&\psi(\mathbf{V}Y^2)&\psi(\mathbf{V}Y)\\
				\psi(\mathbf{V}X)&\psi(\mathbf{V}Y)&\mathbf{I}_{N\times N}\\
			\end{array}} \right],
		\end{equation}
		and
		\begin{equation}
			\mathbf{K} = \left[ {\begin{array}{*{20}{c}}
					\mathbf{V}\psi(X)&\mathbf{0}&\mathbf{0}\\
					\mathbf{0}&\mathbf{V}\psi(Y)&\mathbf{0}\\
					\mathbf{0}&\mathbf{0}&\mathbf{V}\\
				\end{array}} \right],
			\end{equation}
			where $\mathbf{V}$ is an $N\times N$ kernel matrix whose nonzero elements are given by the weights $\upsilon^I_{ij}$, $\psi(\cdot)$ denotes a diagonalizaition operator,
			$X$ and $Y$ denote $N\times 1$ column vectors containing $i_\mathbf{x}$ and $i_\mathbf{y}$, respectively.
			$X^2 = X\circ X$, $Y^2 = Y\circ Y$, and $XY = X\circ Y$, where $\circ$ denotes the Hadamard product.
			
			Since $\upsilon^I_{ij}$ is a normalized bilateral weight, the matrices $\mathbf{U}$ and $\mathbf{K}$ can be efficiently computed using recent EAF algorithms \cite{Gastal11,He10}.
			Furthermore, since $\mu/\lambda\mathbf{I}+\mathbf{U}$ is a block-diagonal matrix,
			$\mathbf{L}^{t}_{\mathbf{x}}$ can be estimated efficiently using a fast sparse matrix solver \cite{Krishnan13}. After optimizing $\mathbf{L}^{t}_{\mathbf{y}}$ in a similar manner, we then have the continuous affine fields
			$\mathbf{L}^{t}$.
			\vspace{-10pt}
			
			\paragraph{Iterative Inference}
			In our filter-based discrete optimization, exhaustively evaluating the raw and aggregated costs for every label $\mathcal{L}$ is still prohibitively time-consuming.
			Thus we utilize the PMF \cite{Lu13} which jointly leverages label cost filtering and fast randomized PatchMatch search in a high dimensional label space. Our discrete optimization differs from the PMF by optimizing the discrete label space with continuous regularization during the iterations, which facilitates convergence and boosts matching performance.
			
			We first decompose an image $I$ into a set of $K$ disjoint segments
			$I=\{S_k,k=1,...,K\}$ and build its set of spatially adjacent segment neighbors.
			Then for each segment $S_k$, two sets of label candidates from the \emph{propagation} and \emph{random search} steps are evaluated for each graph node in scan order.
			In the propagation step, for each segment $S_k$,
			a candidate pixel $i$ is randomly sampled from each neighboring segment, and a set of current best labels $\mathcal{L}_p$ for $i$ is defined by $\{\mathbf{T}_i\}$.
			For these $\mathcal{L}_p$, EAF-based cost aggregation is then performed for the segment $S_k$. In the random search step, a center-biased random search as done in PatchMatch \cite{Barnes10} is performed for the current segment $S_k$, where a sequence of random labels $\mathcal{L}_r$ sampled around the current best label is evaluated. After an iteration of the propagation and random search steps for all segments,
			we apply continuous optimization as described in the preceding section to regularize the discrete affine transformation fields. After each iteration,
			we enlarge $\mu$ such that $\mu \leftarrow c\mu$ with a constant value $1<c\le 2$ to accelerate convergence. \figref{img:3} summarizes our DCTM method, consisting of discrete and continuous optimization, and \figref{img:4} illustrates the convergence of our DCTM method.
			
			To boost matching performance and convergence of our algorithm,
			we apply our method in a coarse-to-fine manner,
			where images $I^f$ are constructed at $F$ image pyramid levels $f=\{1,...,F\}$
			and affine transform fields $\mathbf{T}^f$ are predicted at level $f$.
			Coarser scale results are then used as initialization for the finer levels.
			Algorithm 1 provides a summary of the overall procedure of our DCTM method.\vspace{-3pt}
			\begin{figure*}
				\centering
				\renewcommand{\thesubfigure}{}
				\subfigure[]
				{\includegraphics[width=0.122\linewidth]{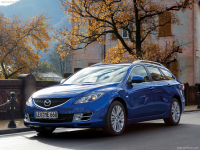}}\hfill
				\subfigure[]
				{\includegraphics[width=0.122\linewidth]{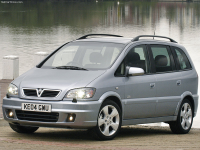}}\hfill
				\subfigure[]
				{\includegraphics[width=0.122\linewidth]{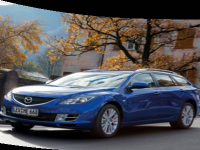}}\hfill
				\subfigure[]
				{\includegraphics[width=0.122\linewidth]{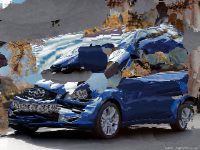}}\hfill
				\subfigure[]
				{\includegraphics[width=0.122\linewidth]{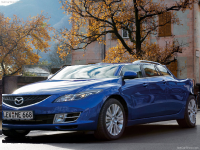}}\hfill
				\subfigure[]
				{\includegraphics[width=0.122\linewidth]{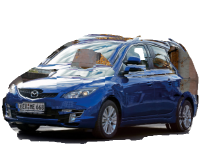}}\hfill
				\subfigure[]
				{\includegraphics[width=0.122\linewidth]{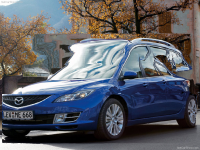}}\hfill
				\subfigure[]
				{\includegraphics[width=0.122\linewidth]{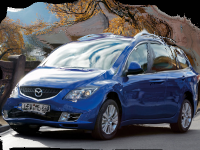}}\hfill
				\vspace{-21.5pt}
				\subfigure[(a)]
				{\includegraphics[width=0.122\linewidth]{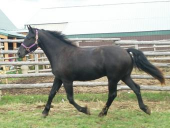}}\hfill
				\subfigure[(b)]
				{\includegraphics[width=0.122\linewidth]{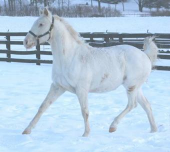}}\hfill
				\subfigure[(c)]
				{\includegraphics[width=0.122\linewidth]{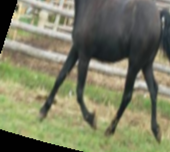}}\hfill
				\subfigure[(d)]
				{\includegraphics[width=0.122\linewidth]{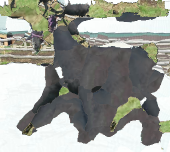}}\hfill
				\subfigure[(e)]
				{\includegraphics[width=0.122\linewidth]{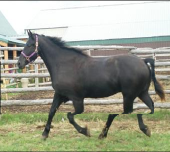}}\hfill
				\subfigure[(f)]
				{\includegraphics[width=0.122\linewidth]{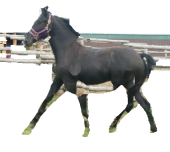}}\hfill
				\subfigure[(g)]
				{\includegraphics[width=0.122\linewidth]{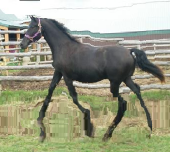}}\hfill
				\subfigure[(h)]
				{\includegraphics[width=0.122\linewidth]{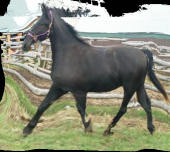}}\hfill
				\vspace{-1pt}
				\caption{Qualitative results on the Taniai benchmark
					\cite{Taniai16}: (a) source image, (b) target image, (c) Lin \emph{et al.} \cite{Lin12}, (d) DFF \cite{Yang14},
					(e) PF \cite{Ham16}, (f) Taniai \emph{et al.} \cite{Taniai16}, (g) SF w/FCSS \cite{Kim17}, and (h) DCTM.
					The source images were warped to the target images using correspondences.}\label{img:5}\vspace{-10pt}
			\end{figure*}
			\begin{figure}
				\centering
				\renewcommand{\thesubfigure}{}
				\subfigure[(a) FG3DCar]
				{\includegraphics[width=0.48\linewidth]{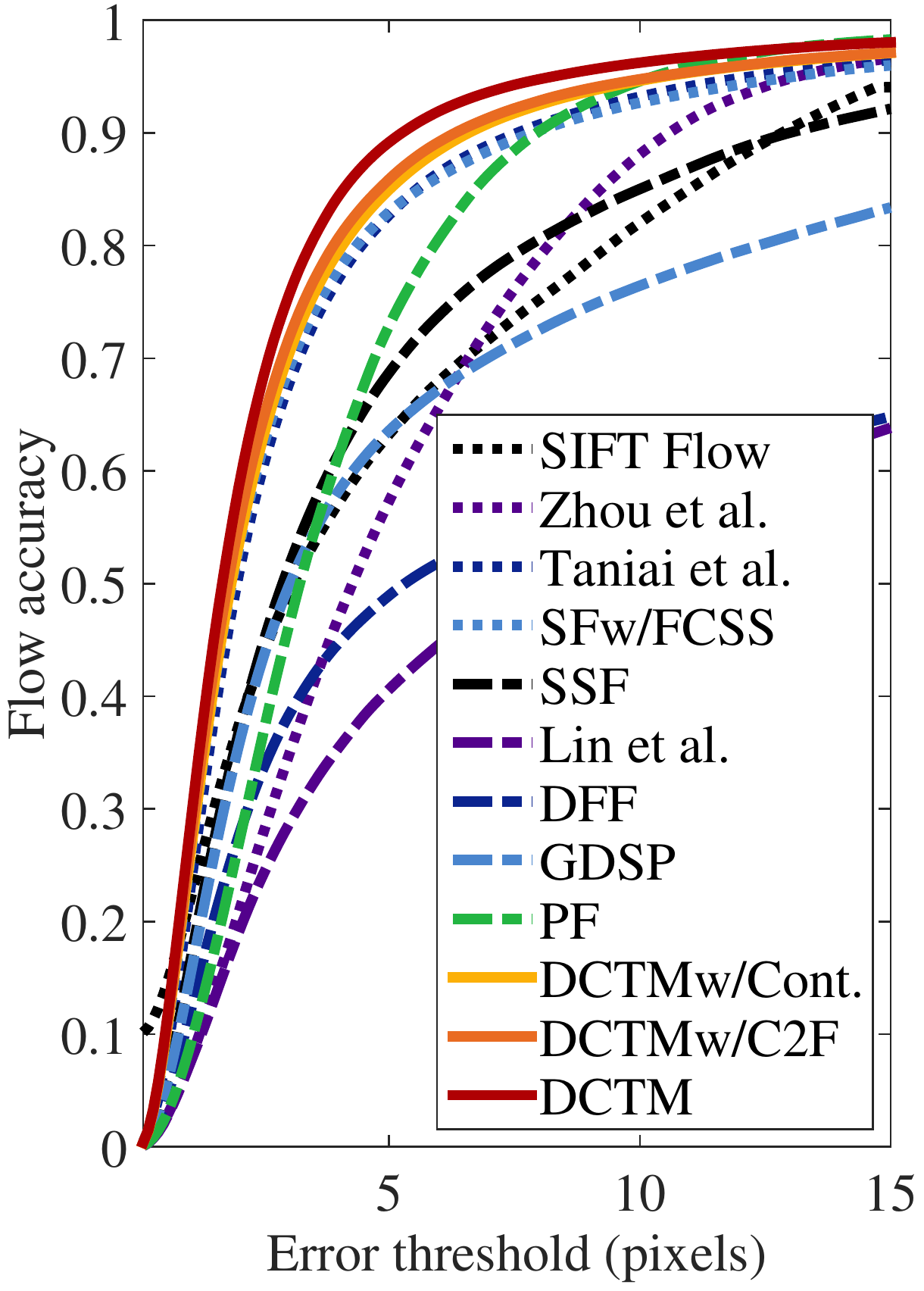}}
				\subfigure[(b) JODS]    
				{\includegraphics[width=0.48\linewidth]{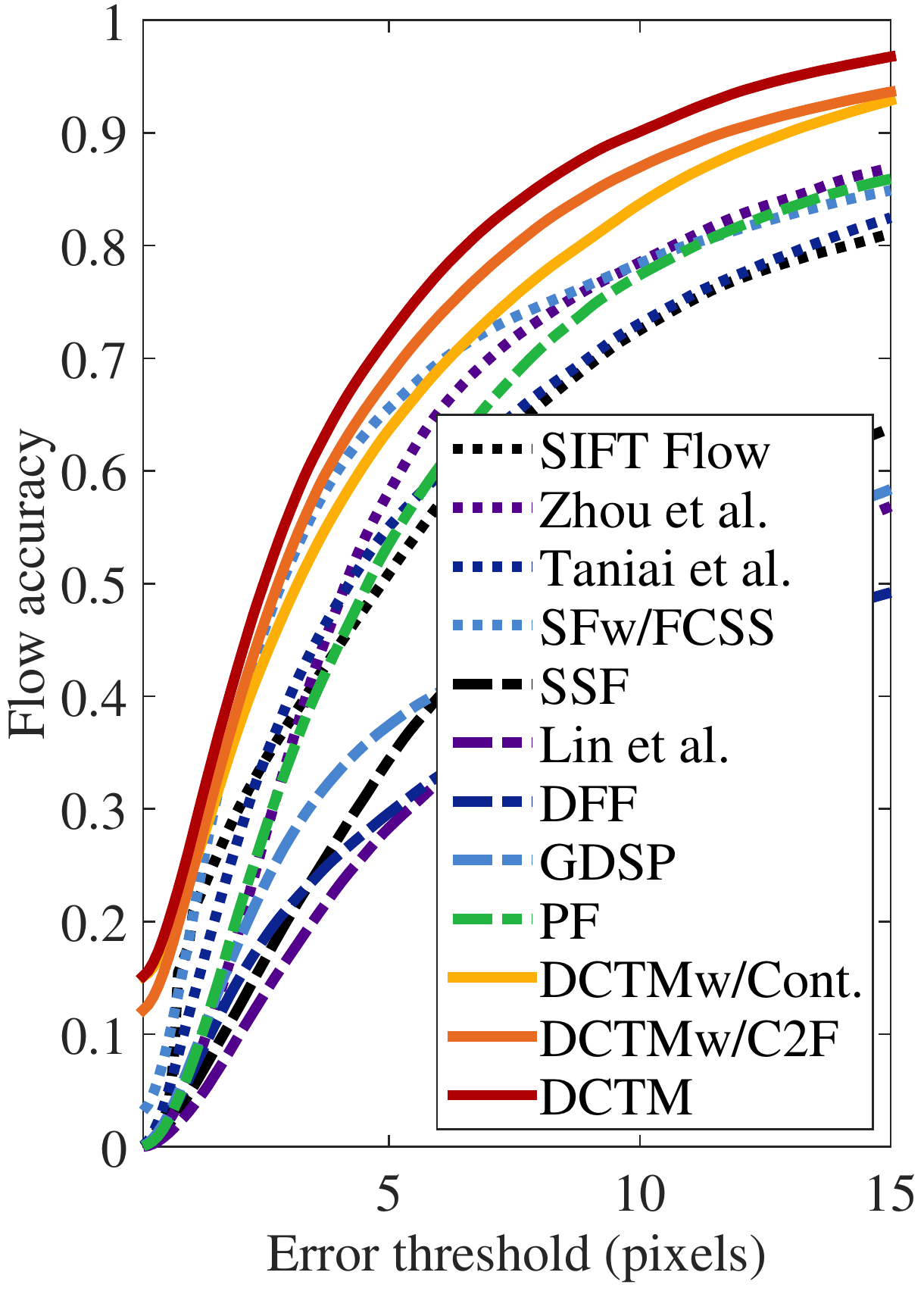}}\\
				\vspace{-6pt}
				\subfigure[(c) PASCAL]  
				{\includegraphics[width=0.48\linewidth]{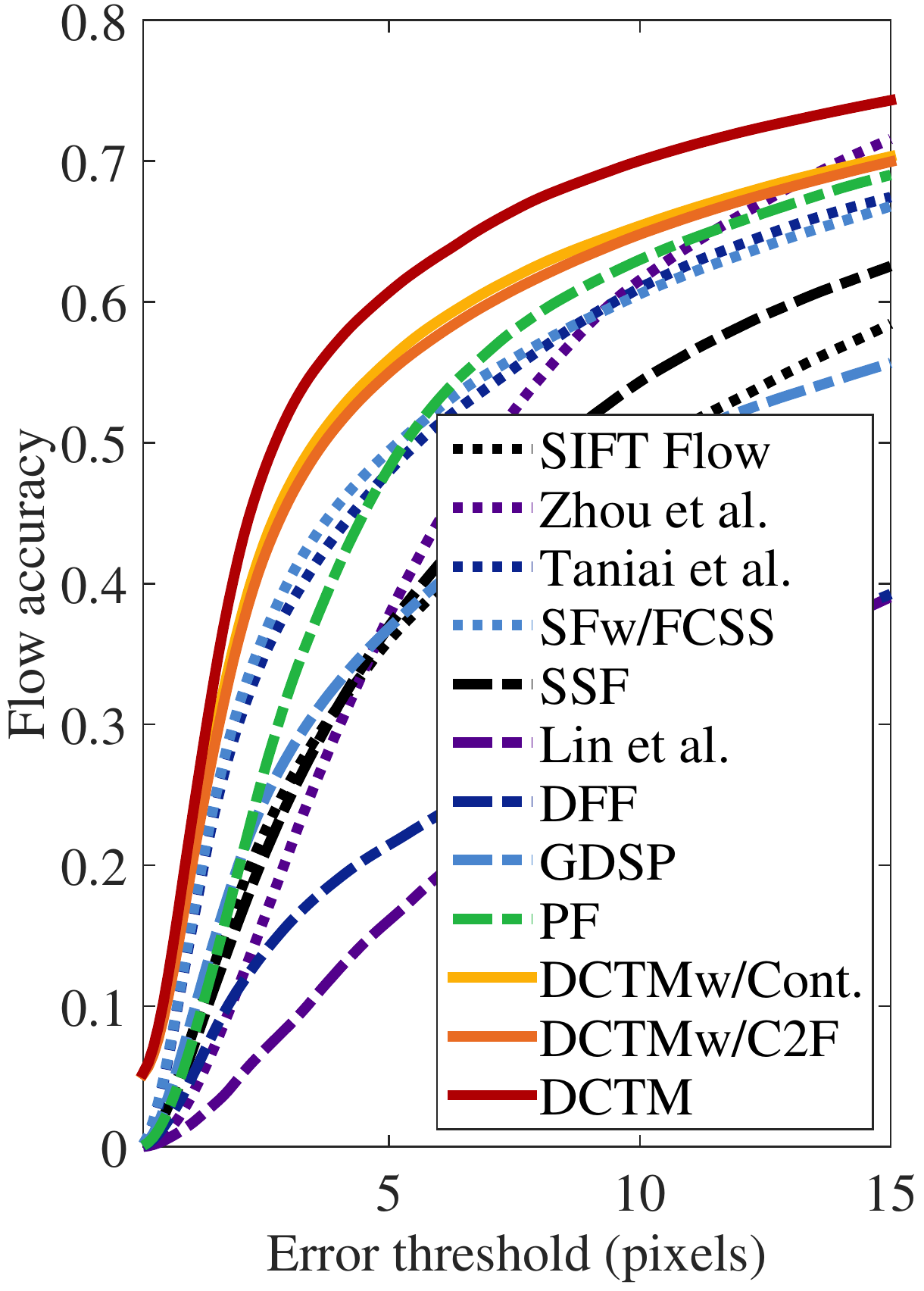}}
				\subfigure[(d) Average] 
				{\includegraphics[width=0.48\linewidth]{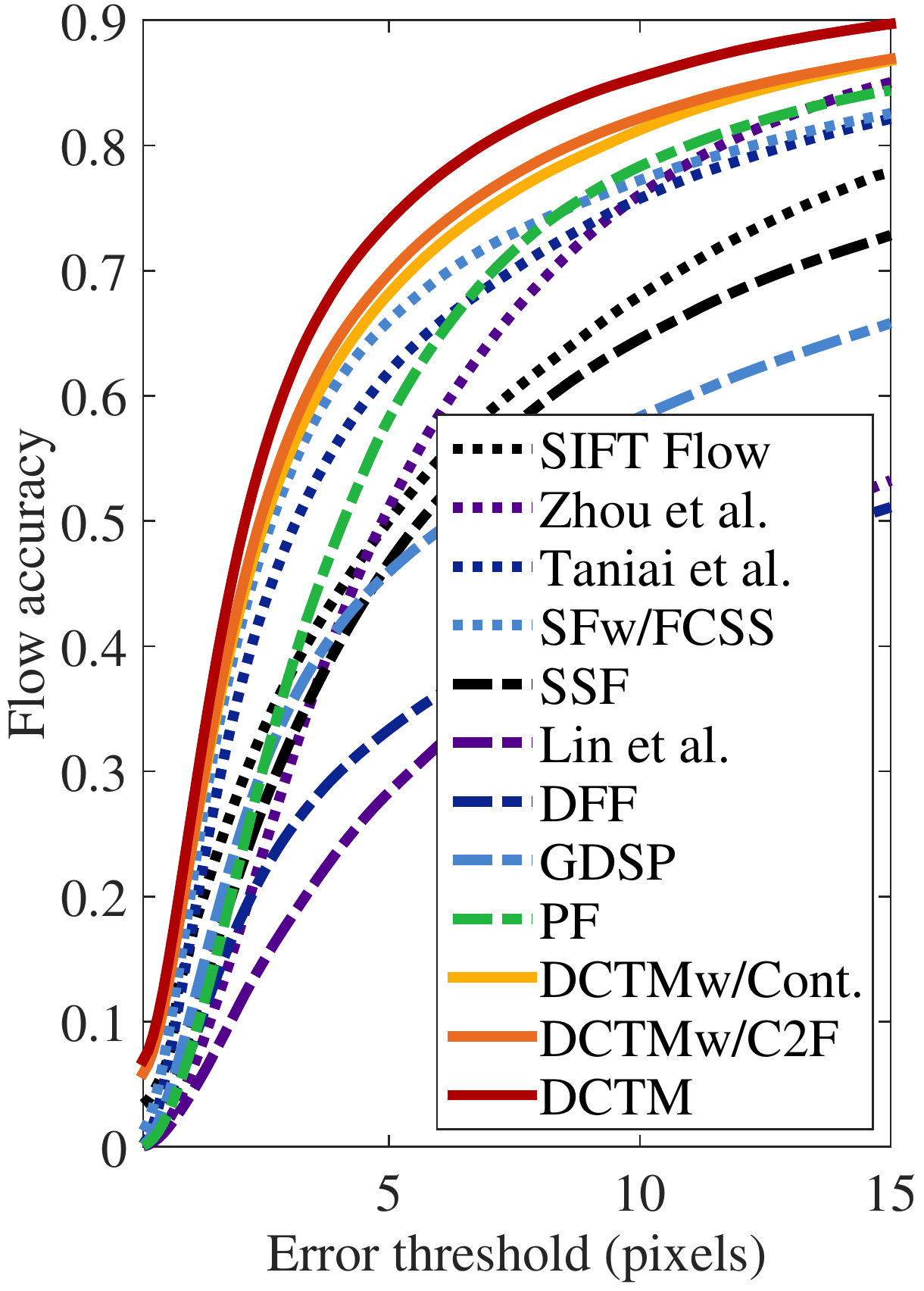}}\\
				\vspace{-1pt}
				\caption{Average flow accuracy with respect to endpoint error threshold on the Taniai benchmark \cite{Taniai16}.
				}\label{img:6}\vspace{-10pt}
			\end{figure}
			\begin{table}[t]
				\centering
				\begin{tabular}{ >{\raggedright}m{0.34\linewidth}
						>{\centering}m{0.10\linewidth} >{\centering}m{0.10\linewidth}
						>{\centering}m{0.10\linewidth} >{\centering}m{0.10\linewidth}}
					\hlinewd{1.0pt}
					Methods &FG3D &JODS &PASC. &Avg.\tabularnewline
					\hline
					\hline
					SIFT Flow \cite{Liu11} &0.632 &0.509 &0.360 &0.500 \tabularnewline
					DSP \cite{Kim13} &0.487 &0.465 &0.382 &0.445\tabularnewline
					Zhou \emph{et al.} \cite{Zhou16} &0.721 &0.514 &0.436 &0.556\tabularnewline
					Taniai \emph{et al.} \cite{Taniai16} &0.830 &0.595 &0.483 &0.636\tabularnewline
					SF w/DAISY \cite{Tola10} &0.636 &0.373 &0.338 &0.449 \tabularnewline
					SF w/VGG \cite{Simonyan15} &0.756 &0.490 &0.360 &0.535 \tabularnewline
					SF w/FCSS \cite{Kim17} &0.830 &0.653 &0.494 &0.660 \tabularnewline
					\hline
					SLS \cite{Hassner12} &0.525 &0.519 &0.320 &0.457 \tabularnewline
					SSF \cite{Qiu14} &0.687 &0.344 &0.370 &0.467 \tabularnewline
					SegSIFT \cite{Trulls13} &0.612 &0.421 &0.331 &0.457 \tabularnewline
					Lin \emph{et al.} \cite{Lin12} &0.406 &0.283 &0.161 &0.283 \tabularnewline
					DFF \cite{Yang14} &0.489 &0.296 &0.214 &0.333 \tabularnewline
					GDSP \cite{Hur15} &0.639 &0.374 &0.368 &0.459 \tabularnewline
					Proposal Flow \cite{Ham16} &0.786 &0.653 &0.531 &0.657 \tabularnewline
					\hline
					DCTM w/DAISY &0.710 &0.506 &0.482 &0.566 \tabularnewline
					DCTM w/VGG &0.790 &0.611 &0.528 &0.630 \tabularnewline
					\hline
					DCTM wo/Cont. &0.850 &0.637 &0.559 &0.682 \tabularnewline
					DCTM wo/C2F &0.859 &0.684 &0.550 &0.698 \tabularnewline
					DCTM &\textbf{0.891} &\textbf{0.721} &\textbf{0.610} &\textbf{0.740} \tabularnewline
					\hlinewd{1.0pt}
				\end{tabular}\vspace{+3pt}
				\caption{Matching accuracy compared to state-of-the-art correspondence techniques on the Taniai benchmark \cite{Taniai16}.}\label{tab:1}\vspace{-10pt}
			\end{table}
			
			\section{Experimental Results}\label{sec:4}
			\subsection{Experimental Settings}\label{sec:41}
			For our experiments, we used the FCSS descriptor provided by authors, which is learned on Caltech-101 dataset \cite{Fei-Fei06}. For EAF for ${\omega^I_{ij}}$ and ${\upsilon^I_{ij}}$, we utilized the guided filter \cite{He13}, where the radius and smoothness parameters are set to $\{16,0.01\}$. The weights in energy function were initially set to $\{\lambda,\mu\}  = \{0.01,0.1\}$ by cross-validation, but $\mu$ increases as evolving iterations with $c=1.8$.
			The SLIC \cite{Achanta12} segment number $K$ increases sublinearly with the image size, \emph{e.g.}, $K=500$ for $640 \times 480$ images.
			The image pyramid level $F$ is set to $3$. We implemented our DCTM method in Matlab/C++ on Intel Core i7-3770 CPU at 3.40 GHz,
			and measured the runtime on a single CPU core. Our code will be made publicly available.
			
			In the following, we comprehensively evaluated our DCTM method through comparisons to the state-of-the-art methods for dense semantic correspondences,
			including SIFT Flow \cite{Liu11}, DSP \cite{Kim13}, Zhou \emph{et al.} \cite{Zhou16}, UCN \cite{Choy16}, Taniai \emph{et al.} \cite{Taniai16}, SIFT Flow optimization with VGG\footnote{In the `VGG', ImageNet pretrained VGG-Net \cite{Simonyan15} from the botton conv1 to the conv3-4 layer were used with $L_2$ normalization \cite{Song16}.} \cite{Simonyan15} and FCSS \cite{Kim17} descriptor.
			Furthermore geometric-invariant methods including SLS \cite{Hassner12}, SSF \cite{Qiu14}, SegSIFT \cite{Trulls13}, Lin \emph{et al.} \cite{Lin12}, DFF \cite{Yang14}, GDSP \cite{Hur15}, and PF \cite{Ham16} were evaluated.
			The performance was measured on Taniai benchmark \cite{Taniai16}, Proposal Flow dataset \cite{Ham16}, and PASCAL-VOC dataset \cite{Chen14}. To validate the components of our method,
			we additionally examined the performance contributions of the continuous optimization (wo/Cont.) and the coarse-to-fine scheme (wo/C2F). Furthermore the performance of our DCTM method when combined with other dense descriptors\footnote{These experiments use only the upright version of the descriptors.} was examined using the DAISY \cite{Tola10} and VGG \cite{Simonyan15}.
			
			\subsection{Results}\label{sec:42}
			\paragraph{Taniai Benchmark \cite{Taniai16}}
			We first evaluated our DCTM method on the Taniai benchmark \cite{Taniai16},
			which consists of 400 image pairs divided into three groups:
			FG3DCar \cite{Lin14}, JODS \cite{Rubinstein13}, and PASCAL \cite{Hariharan11}.
			As in \cite{Taniai16,Kim17}, flow accuracy was
			measured by computing the proportion of foreground pixels with an
			absolute flow endpoint error that is smaller than a certain threshold $T$, after resizing images so that its larger dimension is 100 pixels.
			
			\tabref{tab:1} summarizes the matching accuracy for state-of-the-art correspondence techniques ($T=5$ pixels). \figref{img:5} displays qualitative results for dense flow estimation.
			\figref{img:6} plots the flow accuracy with respect to error threshold. 
			Compared to methods based on handcrafted features \cite{Qiu14,Yang14,Hur15}, CNN based methods \cite{Taniai16,Kim17} provide higher accuracy even though they do not consider geometric variations. The method of Lin \emph{et al.} \cite{Lin12} cannot estimate reliable correspondences due to unstable sparse correspondences. Thanks to its discrete labeling optimization with continuous regularization and affine-FCSS, our DCTM method provides state-of-the-art performance.
			\vspace{-10pt}
			\begin{figure*}[t]
				\centering
				\renewcommand{\thesubfigure}{}
				\subfigure[]
				{\includegraphics[width=0.122\linewidth]{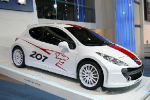}}\hfill
				\subfigure[]
				{\includegraphics[width=0.122\linewidth]{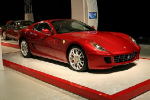}}\hfill
				\subfigure[]
				{\includegraphics[width=0.122\linewidth]{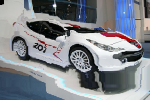}}\hfill
				\subfigure[]
				{\includegraphics[width=0.122\linewidth]{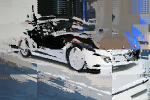}}\hfill
				\subfigure[]
				{\includegraphics[width=0.122\linewidth]{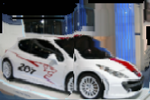}}\hfill
				\subfigure[]
				{\includegraphics[width=0.122\linewidth]{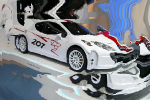}}\hfill
				\subfigure[]
				{\includegraphics[width=0.122\linewidth]{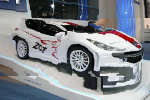}}\hfill
				\subfigure[]
				{\includegraphics[width=0.122\linewidth]{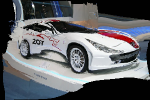}}\hfill
				\vspace{-21.5pt}
				\subfigure[(a)]
				{\includegraphics[width=0.122\linewidth]{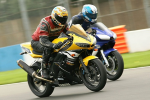}}\hfill
				\subfigure[(b)]
				{\includegraphics[width=0.122\linewidth]{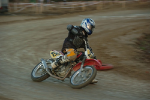}}\hfill
				\subfigure[(c)]
				{\includegraphics[width=0.122\linewidth]{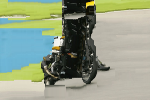}}\hfill
				\subfigure[(d)]
				{\includegraphics[width=0.122\linewidth]{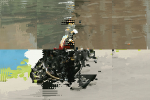}}\hfill
				\subfigure[(e)]
				{\includegraphics[width=0.122\linewidth]{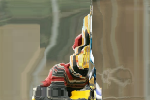}}\hfill
				\subfigure[(f)]
				{\includegraphics[width=0.122\linewidth]{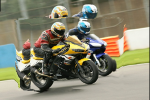}}\hfill
				\subfigure[(g)]
				{\includegraphics[width=0.122\linewidth]{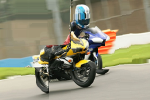}}\hfill
				\subfigure[(h)]
				{\includegraphics[width=0.122\linewidth]{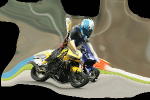}}\hfill
				\vspace{-1pt}
				\caption{Qualitative results on the Proposal Flow benchmark \cite{Ham16}: (a) source image, (b) target image,
					(c) SSF \cite{Qiu14}, (d) DSP \cite{Kim13}, (e) GDSP \cite{Hur15}, (f) PF \cite{Ham16}, (g) SF w/FCSS \cite{Kim17}, and (h) DCTM.
					The source images were warped to the target images using correspondences.}\label{img:7}\vspace{-10pt}
			\end{figure*}
			\begin{figure*}[t]
				\centering
				\renewcommand{\thesubfigure}{}
				\subfigure[]
				{\includegraphics[width=0.108\linewidth]{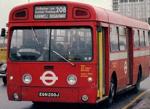}}\hfill
				\subfigure[]
				{\includegraphics[width=0.108\linewidth]{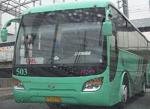}}\hfill
				\subfigure[]
				{\includegraphics[width=0.108\linewidth]{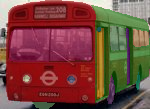}}\hfill
				\subfigure[]
				{\includegraphics[width=0.108\linewidth]{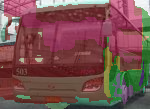}}\hfill
				\subfigure[]
				{\includegraphics[width=0.108\linewidth]{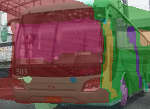}}\hfill
				\subfigure[]
				{\includegraphics[width=0.108\linewidth]{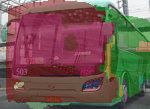}}\hfill
				\subfigure[]
				{\includegraphics[width=0.108\linewidth]{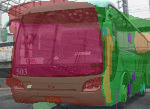}}\hfill
				\subfigure[]
				{\includegraphics[width=0.108\linewidth]{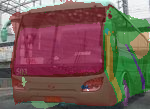}}\hfill
				\subfigure[]
				{\includegraphics[width=0.108\linewidth]{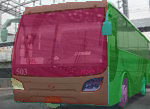}}\hfill    
				\vspace{-21.5pt}
				\subfigure[(a)]
				{\includegraphics[width=0.108\linewidth]{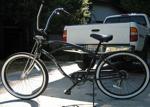}}\hfill
				\subfigure[(b)]
				{\includegraphics[width=0.108\linewidth]{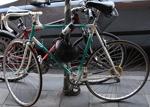}}\hfill
				\subfigure[(c)]
				{\includegraphics[width=0.108\linewidth]{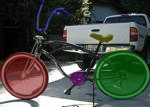}}\hfill
				\subfigure[(d)]
				{\includegraphics[width=0.108\linewidth]{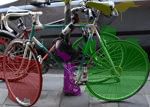}}\hfill
				\subfigure[(e)]
				{\includegraphics[width=0.108\linewidth]{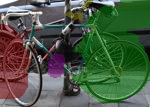}}\hfill
				\subfigure[(f)]
				{\includegraphics[width=0.108\linewidth]{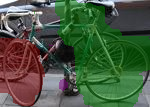}}\hfill
				\subfigure[(g)]
				{\includegraphics[width=0.108\linewidth]{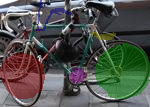}}\hfill
				\subfigure[(h)]
				{\includegraphics[width=0.108\linewidth]{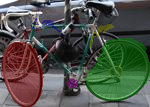}}\hfill
				\subfigure[(i)]
				{\includegraphics[width=0.108\linewidth]{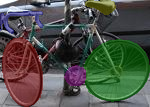}}\hfill
				\vspace{-1pt}
				\caption{Visualizations of dense flow field with color-coded part segments on the PASCAL-VOC part dataset \cite{Chen14}: (a) source image, (b) target image, (c) source mask, (d) DFF \cite{Yang14}, (e) GDSP \cite{Hur15}, (f) Zhou \emph{et al.} \cite{Zhou16}, (g) SF w/FCSS \cite{Kim17}, (h) DCTM, and (i) target mask.}\label{img:8}\vspace{-10pt}
			\end{figure*}
			\begin{table}[t]
				\centering
				\begin{tabular}{ >{\raggedright}m{0.33\linewidth}
						>{\centering}m{0.16\linewidth}  >{\centering}m{0.14\linewidth}
						>{\centering}m{0.16\linewidth}}
					\hlinewd{1.0pt}
					\multirow{2}{*}{Methods} &\multicolumn{3}{ c }{PCK} \tabularnewline
					\cline{2-4}
					&$\alpha=0.05$ &$\alpha=0.1$ &$\alpha=0.15$\tabularnewline
					\hline
					\hline
					SIFT Flow \cite{Liu11} &0.247 &0.380 &0.504 \tabularnewline
					DSP \cite{Kim13} &0.239 &0.364 &0.493 \tabularnewline
					Zhou \emph{et al.} \cite{Zhou16} &0.197 &0.524 &0.664 \tabularnewline
					SF w/FCSS \cite{Kim17} &0.354 &0.532 &0.681 \tabularnewline
					\hline 
					SSF \cite{Qiu14} &0.292 &0.401 &0.531 \tabularnewline
					Lin \emph{et al.} \cite{Lin12} &0.192 &0.354 &0.487 \tabularnewline
					DFF \cite{Yang14} &0.241 &0.362 &0.510 \tabularnewline
					GDSP \cite{Hur15} &0.242 &0.487 &0.512 \tabularnewline
					Proposal Flow \cite{Ham16} &0.284 &0.568 &0.682 \tabularnewline		
					\hline
					DCTM &\textbf{0.381} &\textbf{0.610} &\textbf{0.721} \tabularnewline
					\hlinewd{1.0pt}
				\end{tabular}\vspace{+3pt}
				\caption{Matching accuracy compared to state-of-the-art correspondence techniques on the Proposal Flow benchmark \cite{Ham16}.}\label{tab:2}\vspace{-10pt}
			\end{table}
			
			\paragraph{Proposal Flow Benchmark \cite{Ham16}}
			We also evaluated our FCSS descriptor on the Proposal Flow benchmark \cite{Ham16},
			which includes 10 object sub-classes with 10 keypoint annotations for each image.
			For the evaluation metric, we used the probability of correct keypoint (PCK) between flow-warped keypoints and the ground truth \cite{Long14,Ham16}. The warped keypoints are deemed to be correctly predicted if they lie within
			$\alpha \cdot \mathrm{max}(H,W)$ pixels of the ground-truth keypoints for
			$\alpha \in [0,1]$, where $H$ and $W$ are the height and width of
			the object bounding box, respectively. The PCK values were measured for
			different correspondence techniques in \tabref{tab:2}.
			\figref{img:7} shows qualitative results for dense flow estimation. Our DCTM method exhibits performance competitive to the state-of-the-art correspondence techniques.
			\vspace{-10pt}
			\paragraph{PASCAL-VOC Parts Dataset \cite{Chen14}}
			Lastly, we evaluated our DCTM method on the dataset provided by \cite{Zhou15}, where the images are sampled from the PASCAL parts dataset \cite{Chen14}. With human-annotated part segments, we measured part matching accuracy using the weighted intersection over union (IoU) score between transferred segments and ground truths, with weights determined by the pixel area of each part. To evaluate alignment accuracy, we measured the PCK metric using keypoint annotations for the 12 rigid PASCAL classes \cite{Xiang14}. \tabref{tab:3} summarizes the matching accuracy compared to state-of-the-art correspondence methods. \figref{img:8} visualizes estimated dense flow with color-coded part segments. From the results, our DCTM method is found to yield the highest matching accuracy.
			\vspace{-10pt}
			\paragraph{Computation Speed}
			For all the test cases, our DCTM method converges with $3$-$5$ iterations on each image pyramid level. For $320 \times 240$ images, the average runtime of DCTM is $15$-$20$ seconds, compared to $216$ seconds for GDSP \cite{Hur15}, $73$ seconds for DFF \cite{Yang14}, $276$ seconds for Lin \emph{et al.} \cite{Lin12}, and $321$ seconds for Taniai \emph{et al.} \cite{Taniai16}.\vspace{-3pt}
			\begin{table}[t]
				\centering
				\begin{tabular}{ >{\raggedright}m{0.33\linewidth}
						>{\centering}m{0.08\linewidth}  >{\centering}m{0.16\linewidth} >{\centering}m{0.14\linewidth}}
					\hlinewd{1.0pt}
					\multirow{2}{*}{Methods} &\multirow{2}{*}{IoU} &\multicolumn{2}{ c }{PCK} \tabularnewline
					\cline{3-4}
					& &$\alpha=0.05$ &$\alpha=0.1$\tabularnewline
					\hline
					\hline
					Zhou \emph{et al.} \cite{Zhou16} &- &- &0.24 \tabularnewline
					UCN \cite{Choy16} &- &0.26 &0.44\tabularnewline
					SF w/ FCSS \cite{Liu11} &0.44 &0.28 &0.47\tabularnewline
					\hline
					DFF \cite{Yang14} &0.36 &0.14 &0.31 \tabularnewline
					GDSP \cite{Hur15} &0.40 &0.16 &0.34 \tabularnewline
					Proposal Flow \cite{Ham16} &0.41 &0.17 &0.36\tabularnewline
					\hline
					DCTM &\textbf{0.48} &\textbf{0.32} &\textbf{0.50}\tabularnewline
					\hlinewd{1.0pt}
				\end{tabular}\vspace{+3pt}
				\caption{Matching accuracy on the PASCAL-VOC dataset \cite{Chen14}.}\label{tab:3}\vspace{-10pt}
			\end{table}
			
			\section{Conclusion}\label{sec:5}
			We presented DCTM, which estimates dense affine transformation fields through a discrete label optimization in which the labels are iteratively updated via continuous regularization. DCTM infers solutions from the continuous space of affine transformations in a manner that can be computed efficiently through constant-time edge-aware filtering and the affine-FCSS descriptor. A direction for further study is to examine how the semantic flow of DCTM could benefit single-image 3D reconstruction and instance-level object segmentation.
			
			{\small
				\bibliographystyle{ieee}
				\bibliography{egbib}
			}

\end{document}